\documentclass[conference]{IEEEtran}
\usepackage{graphics} 
\usepackage{epsfig} 
\usepackage{mathptmx} 
\usepackage{amsmath} 
\usepackage{amssymb}  
\usepackage{graphicx}
\usepackage{eucal}
\usepackage{todonotes}[inline]  

\usepackage{float}
\usepackage[ruled,vlined]{algorithm2e}
\usepackage{pgfplotstable}
\usepackage{pgfplots}
\usepackage{tikz}
\usetikzlibrary[pgfplots.groupplots]
\usepgfplotslibrary{fillbetween}
\pgfplotsset{compat=1.5}
\usepackage{subcaption}
\usepackage{caption}
\usepackage{xcolor}
\usepackage{enumerate}
\usepackage{makecell}
\usepackage{colortbl}

\usepackage{graphicx}
\usepackage{tabularx,ragged2e,booktabs}
\usepackage{url}
\usepackage{siunitx}
\usepackage{rotating}
\usepackage{amssymb}
\usepackage{tabularx}
\usepackage{pgfplots}
\usepackage{csvsimple}
\usepackage{multirow}
\usepackage{cite}
\usepackage{float}
\usepackage{graphicx}
\usepackage{booktabs}

\ifCLASSINFOpdf
\else
\fi
\DeclareMathOperator*{\suppremum}{sup} 

\begin{document}
%
\title{Towards Robust 3D Object Recognition with \\ Dense-to-Sparse Deep Domain Adaptation }



%
\author{\IEEEauthorblockN{Prajval Kumar Murali\IEEEauthorrefmark{1}\IEEEauthorrefmark{3},
Cong Wang\IEEEauthorrefmark{1}\IEEEauthorrefmark{2},
Ravinder Dahiya\IEEEauthorrefmark{3}, and
Mohsen Kaboli\IEEEauthorrefmark{1}\IEEEauthorrefmark{4}}
\IEEEauthorblockA{\IEEEauthorrefmark{1}BMW Group, M\"unchen, Germany, \IEEEauthorrefmark{2}Technical University of Munich, Germany.}
\IEEEauthorblockA{\IEEEauthorrefmark{3}University of Glasgow, Scotland, \IEEEauthorrefmark{4}Radboud University, Netherlands. \\Email: prajval-kumar.murali@bmwgroup.com, ge57qom@mytum.de,\\ ravinder.dahiya@glasgow.ac.uk, mohsen.kaboli@bmwgroup.com}}

\maketitle
\begin{abstract}
Three-dimensional (3D) object recognition is crucial for intelligent autonomous agents such as autonomous vehicles and robots alike to operate effectively in unstructured environments. Most state-of-art approaches rely on relatively dense point clouds and performance drops significantly for sparse point clouds. Unsupervised domain adaption allows to minimise the discrepancy between dense and sparse point clouds with minimal unlabelled sparse point clouds, thereby saving additional sparse data collection, annotation and retraining costs. In this work, we propose a novel method for point cloud based object recognition with competitive performance with state-of-art methods on dense and sparse point clouds while being trained only with dense point clouds.
\end{abstract}

%
\IEEEpeerreviewmaketitle

\section{Introduction}
\label{sec:introduction}
\begin{figure}[t!]
    \centering
    \includegraphics[width = 0.8\columnwidth]{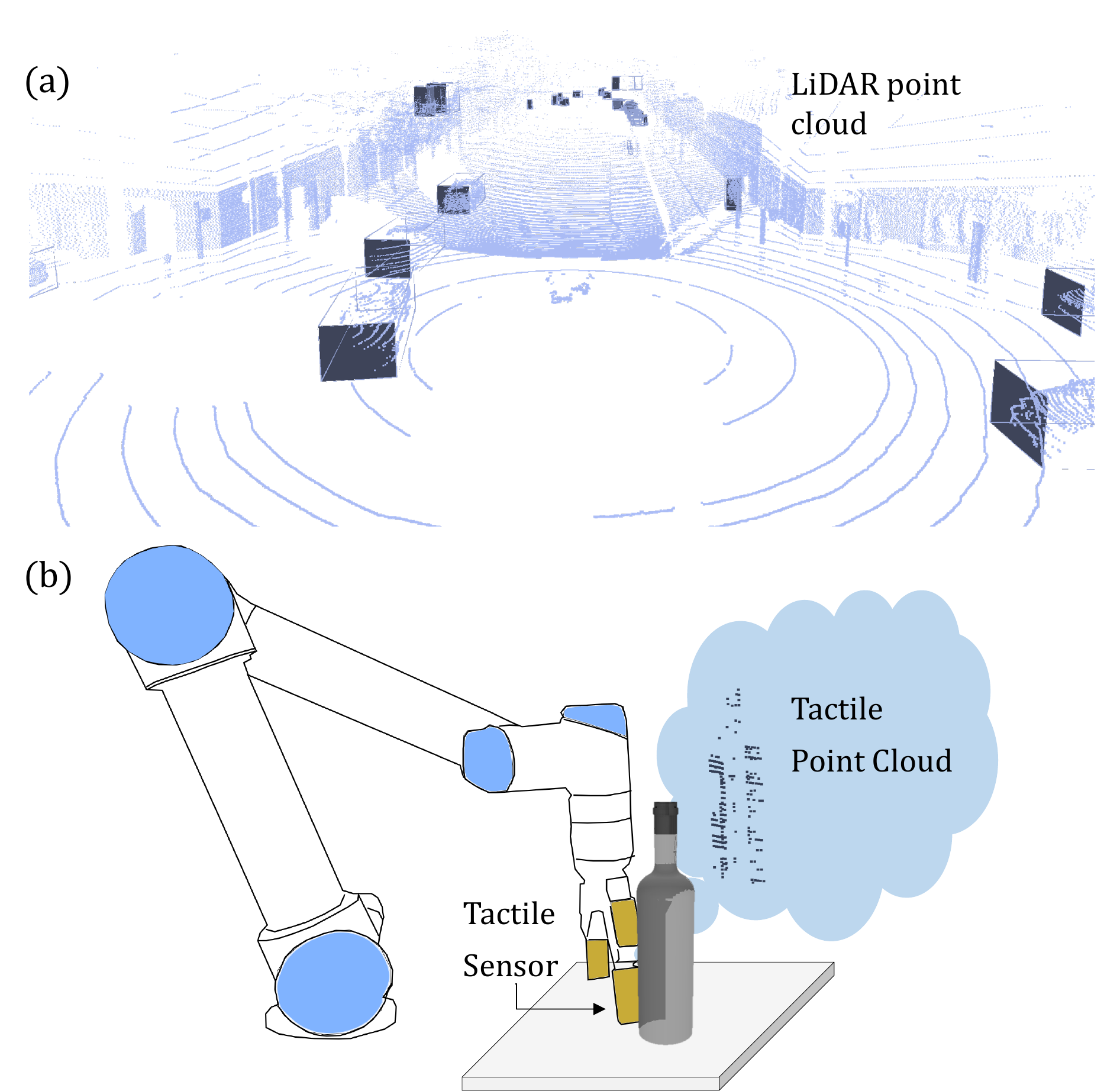}
    \caption{Typical applications with sparse point clouds: (a) LiDAR scans in autonomous vehicles for object/ pedestrian detection and localisation (from PandaSet~\cite{xiao2021pandaset}); (b) robots with tactile sensors performing object recognition and manipulation.}
    \label{fig:fig1}
\end{figure}

Point cloud based approaches are becoming popular for various applications in the automotive and robotics domain such as object recognition, scene understanding, pose estimation and so on due to the ubiquity of 3D visual sensors. 
Typical visual sensors such as RGB-D cameras produce around $10^4$ points with a single snapshot. However, LiDAR-based point clouds are very sparse with increasing distance, ranging between 10-100 points~\cite{wang2021sparse, murali2021intelligent}. 
Such sparse point clouds are also produced with commercial tactile sensors~\cite{Qiang-TRO-2020,murali2021active, murali2022active} (Figure~\ref{fig:fig1}). 
Most state-of-art methodologies rely on relatively dense point clouds and the performance drops drastically with sparse point clouds~\cite{xiao2021triangle}. In particular case for tactile sensing, acquiring larger number of point clouds is directly proportional to the number of contact actions performed using the tactile sensor~\cite{kaboli2019tactile, kaboli2018robust, kaboli2018active, kaboli2016tactilehuman, kaboli2017tactile, murali2022infogain}. Hence, for increased robustness and efficiency of autonomous agents (robots or autonomous vehicles) it is imperative for perception algorithms to perform adequately with limited amount of data availability. 

Recently, deep learning approaches have been proposed for working with raw 3D point clouds for various tasks such as semantic segmentation, object pose estimation and object recognition~\cite{guo2020deep}. 
However, it is challenging due to the unstructured nature of point clouds, small-scale datasets and high dimensionality of the data~\cite{guo2020deep}.
PointNet~\cite{qi2017pointnet} and PointNet++~\cite{qi2017pointnet++} were the first works in this domain and several derivative networks have been proposed based on the PointNet architecture~\cite{guo2020deep}. According to various studies, the performance of such networks degrades dramatically when sparse point clouds with point numbers ranging from 10 to 100 are used~\cite{qi2017pointnet++, xiao2021triangle, stutz2020learning, liu2019relation}. 
The sparse point clouds pose a significant challenge due to inability to extract meaningful features such as surface normals and local curvature. Even for humans, it becomes increasingly difficult to discern the point clouds for identifying objects when the number of points falls below 100~\cite{xiao2021triangle}.
Using the ModelNet40 dataset~\cite{wu20153d}, Xiao et al.~\cite{xiao2021triangle} introduced Triangle-Net which is trained on dense and sparse point clouds and demonstrates a classification accuracy of 70\% using less than 20 points.
However, retraining large networks with sparse point cloud data adds additional overhead costs such as data acquisition, data labelling and training. Domain adaptation techniques can help bridge the gap by using the trained networks on dense point clouds that are relatively easily available with a smaller subset of unlabelled sparse point clouds.
This helps reduce data collection and labelling costs by a human annotator while increasing model robustness by allowing the models to work on various types of input data.
There are several approaches for unsupervised domain adaptation available in literature~\cite{wilson2020survey, wang2018deep} wherein we concentrate on discrepancy based techniques such as Maximum mean discrepancy (MMD)~\cite{borgwardt2006integrating} and Correlation Alignment (CORAL)~\cite{sun2016return}. The discrepancy-based methods works by reducing the distance between the source and target domains using statistic criteria.  
\newline \textbf{Contributions:}
\newline In this work, we propose a novel unsupervised domain adaptation method from dense to sparse point clouds for object recognition. Our network is trained with dense labelled point clouds and adapted with a small subset of unlabelled sparse point clouds. We use the ModelNet10~\cite{wu20153d} dataset for evaluating our proposed method. Furthermore, we benchmark our approach against other state-of-art point cloud classification methods such as Triangle-Net~\cite{xiao2021triangle}, PointNet~\cite{qi2017pointnet}, and DGCNN~\cite{zhang2018end}.

\section{Methodology}
\label{sec:methods}
\begin{figure}[t!]
    \centering
    \includegraphics[width = 0.9\columnwidth]{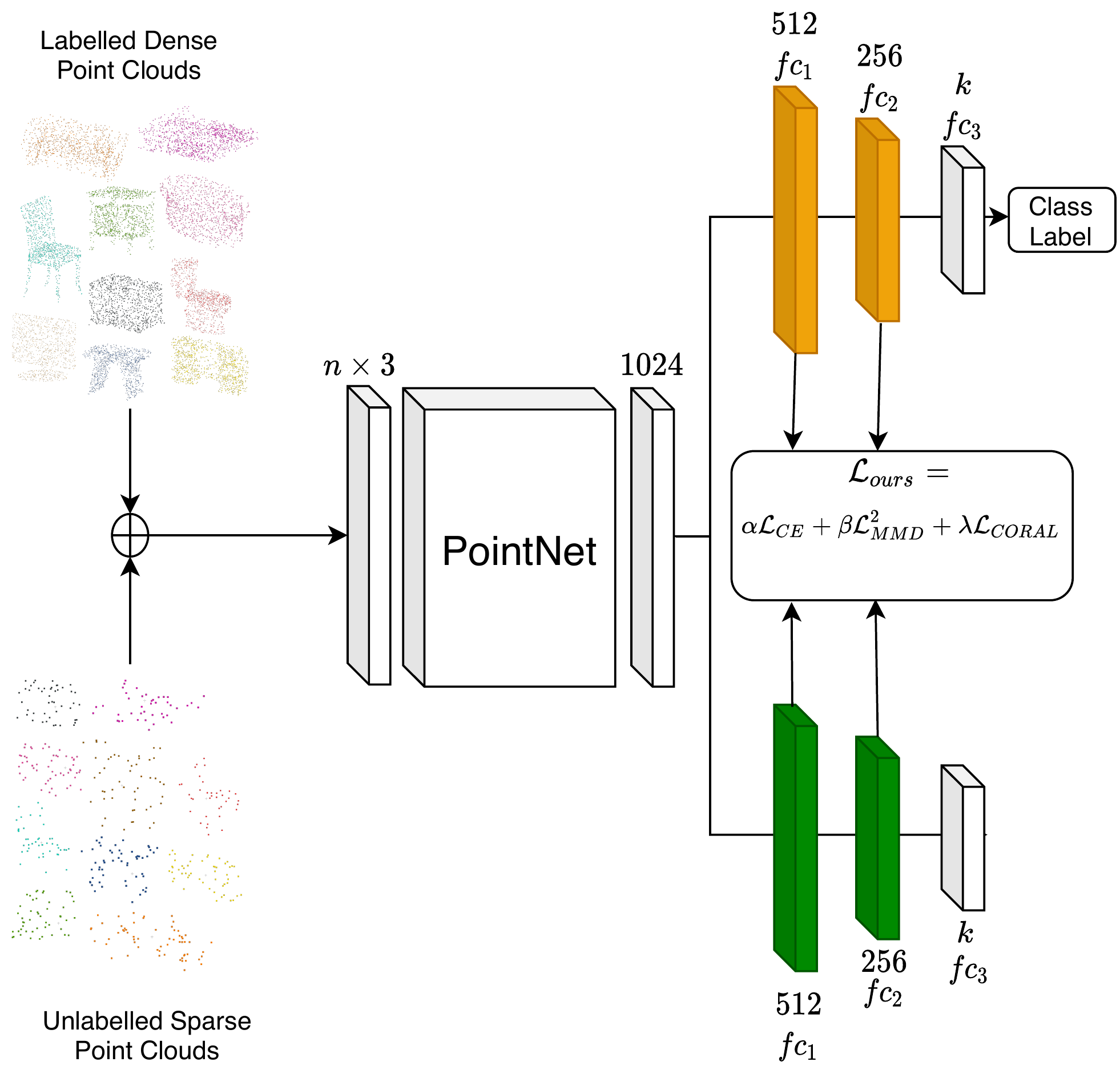}
    \caption{Network architecture for dense to sparse unsupervised domain adaptation for object recognition}
    \label{fig:network_arch}
\end{figure}
As the target domain has no labelled data, it is not possible to fine-tune the network that is trained on the source domain data to the target domain. Hence, we train our network on the labelled source dataset and use discrepancy methods of domain adaptation to adapt the trained model with the unlabelled target dataset. Our network architecture is shown in Figure~\ref{fig:network_arch}. The input to the network is $n$ point clouds with $x,y,z$ data associated with each point. The PointNet~\cite{qi2017pointnet} architecture is used for extracting the features from the raw point cloud data. The network provides as output $k$ classification scores for each of the $k$ classes. We get a global feature vector of size 1024 from the PointNet output which is used by a series of fully connected layers for the classification task. We use 3 fully connected layers $fc$ of sizes $512, 256, k$. 
We trained our network with dense point clouds from ModelNet10~\cite{wu20153d} wherein the point clouds of size 1024 points are sampled randomly from the CAD meshes in ModelNet10~\cite{wu20153d}.
The output size is $k=10$ as there are 10 classes of objects in ModelNet10~\cite{wu20153d}.
We use the cross-entropy loss for training. This represents the source dataset $D_s = \{(x^s_i,y_i^s)\}_{i=1}^{n_s}$ with $n_s$ labelled dense point cloud samples. The target domain is represented as $D_t = \{x^t_j\}_{j=1}^{n_t}$ with $n_t$ unlabelled samples comprising of sparse point clouds of 50 points sampled from the ModelNet10 dataset. We employ discrepancy-based methods to extract domain-invariant representations for the unsupervised domain adaptation. Maximum Mean Discrepancy (MMD)~\cite{borgwardt2006integrating} and Correlation Alignment (CORAL)~\cite{sun2016return} are well-known methods for domain adaptation that has been extensively applied in the visual domain. 
Given labelled source domain $D_s = \{(x^s_i,y_i^s)\}_{i=1}^{n_s}$ and unlabelled target domain $D_t = \{x^t_j\}_{j=1}^{n_t}$ with probability distributions $p^s$ and $p^t$ respectively, the MMD between $p^s$ and $p^t$ is defined as:
\begin{equation}
    \text{MMD}^2(p^s, p^t) = \suppremum_{||\phi||_{\mathcal{H}} \leq 1}||\mathbb{E}_{x^s \sim p^s}[\phi(x^s)] - \mathbb{E}_{x^t \sim p^t}[\phi(x^t)] ||^2_{\mathcal{H}},
    \label{eq:mmd}
\end{equation}
where $\mathcal{H}$ is the reproducing kernel Hilbert space (RKHS), $\phi(.)$ is the feature mapping associated with the kernel map $k(x^s, x^t) = <\phi(x^s), \phi(x^t)>$, $\suppremum(.)$ is the supremum of the input aggregate and $||\phi||_{\mathcal{H}}\leq 1$ defines a set of functions in the unit ball of $\mathcal{H}$. The multi-kernel MMD (MK-MMD)~\cite{gretton2012optimal} obtains the optimal kernel as a linear combination of individual kernels. The kernel $k(x^s, x^t)$ is defined as the convex combination of $b$ positive semi-definite kernels $\{k_u\}$~\cite{gretton2012optimal}:
\begin{equation}
    K \triangleq \left\{ k = \sum_{u=1}^b \beta_u k_u : \sum_{u=1}^b\beta_u=1, \beta_u \geq 0, \forall u\right\},
    \label{eq:kernels}
\end{equation}
wherein $k$ is weighted by the different kernels and the coefficients $\beta_u$ are the weights to ensure that the generated
multi-kernel $k$ is characteristic.
MK-MMD compares all orders of statistics between the source and target domain while in comparison, CORAL~\cite{sun2016return} compares the $2^{nd}$ order statistics of the source and target distributions. Sun et al.~\cite{sun2016deep} proposed Deep-CORAL that uses CORAL for deep neural networks and is defined as follows:
\begin{equation}
    \text{CORAL}(x^s,x^t) = \frac{1}{4d^2}||C_s - C_t||^2_F,
    \label{eq:coral}
\end{equation}
where $||.||^2_F$ is the squared matrix Frobenius norm, $C_s$ and $C_t$ are the covariance matrices of the source and target domain data. 
Our loss function is defined as the weighted linear combination of MK-MMD, CORAL and the classification loss for the source dataset.
Furthermore, previous research has demonstrated that modifying a single layer alone is insufficient to correct the dataset bias between the source and target domains caused by non-transferable layers~\cite{yosinski2014transferable}.
Hence, we also use fully-connected layers $fc_1$ and $fc_2$ for multi-layer domain adaptation.
Our proposed loss function $\mathcal{L}_{ours}$ is:
\begin{equation}
\begin{split}
    \mathcal{L}_{ours}  & = \alpha \mathcal{L}_{crossEnt} + \beta\{\mathcal{L}_{MK-MMD}^2\}_{fc_1} + \beta\{\mathcal{L}_{MK-MMD}^2\}_{fc_2} \\
     & +\lambda\{\mathcal{L}_{CORAL}\}_{fc_1} +\lambda\{\mathcal{L}_{CORAL}\}_{fc_2}, 
    \end{split}
    \label{eq:vtloss}
\end{equation}
where $\alpha, \beta, \lambda$ are hyperparameters that need to be tuned empirically. 
The model is adapted to the target domain by minimising the loss $\mathcal{L}_{ours}$. 

\section{Experimental Results}
\label{sec:experiment}
We used the ModelNet10 dataset containing 4,899 pre-aligned shapes from 10 categories. 
The training and domain adaptation of the network was performed using PyTorch framework on a workstation with NVidia Quadro RTX 4000 GPU with 8 GB RAM. The training set consists of two parts: supervised training with labelled dense point clouds and unsupervised domain adaptation with unlabelled sparse point clouds. The labelled dense point clouds are generated by randomly subsampling 3,991 (80\%) objects in ModelNet10 to 1024 points. The unlabeled sparse point clouds are generated by randomly subsampling the same samples with random reshuffling to 50 points. The test set is generated by randomly subsampling 908 (20\%) objects to 50 points. 
The dataset is further augmented by randomly rotating each sample around the \textit{z} axis for rotation invariance.
The network is first trained by labeled dense point clouds with 150 epochs. Then, the PointNet layers are frozen and the network is trained based on $\mathcal{L}_{ours}$ in Equation(\ref{eq:vtloss}) with unlabelled sparse data.
Table \ref{tab:comparison} shows the overall results against other state-of-art methods such as Triangle-Net~\cite{xiao2021triangle}, PointNet~\cite{qi2017pointnet}, and DGCNN~\cite{zhang2018end}. 
The other state-of-art methods in Table~\ref{tab:comparison} are solely trained with dense point cloud dataset without domain adaptation. 
Table \ref{tab:test_loss} shows the accuracy on the test set when using $\mathcal{L}_{MMD}$, $\mathcal{L}_{CORAL}$ and the $\mathcal{L}_{ours}$ loss in training as ablation studies for different number of points.
Table \ref{tab:test_alpha} shows test accuracy results over different $\alpha$, and Table \ref{tab:test_beta} shows the results over different $\beta$ for hyper-parameter tuning.
\begin{table}[t!]
\caption{Comparison results against state-of-art methods}
\label{tab:comparison}
\resizebox{\columnwidth}{!}{%
\begin{tabular}{cccccccccc}
\toprule
\textbf{Points Num} & 1024     & 50     & 40     & 30     & 20     & 10  \\
\midrule
\textbf{PointNet}~\cite{qi2017pointnet}    & 92.32\%  & 81.03\% & 76.86\% & 68.09\% & 58.55\% & 32.89\% \\
\textbf{TriangleNet}~\cite{xiao2021triangle}            & 81.36\%  & 64.45\% & 59.89\% & 47.97\% & 33.79\% & 11.06\% \\
\textbf{DGCNN}~\cite{zhang2018end}      & 93.27\%  & 3.69\% & 3.12\% & 2.03\% & 0.89\% & N.A \\
\textit{\textbf{Our Model}} & 91.12\%  & 82.35\% & 81.69\% & 74.89\% & 65.58\% & 41.18\% \\
\bottomrule
\end{tabular}%
}
\end{table}

\begin{table}[t!]
\centering
\caption{Test Result on ModelNet10}
\label{tab:test_loss}
\begin{tabular}{ccccc}
\toprule
\textbf{Points Num}  & $\mathcal{L}_{CORAL}$   & $\mathcal{L}_{MMD}$ & $\mathcal{L}_{ours}$\\
\midrule
1024         & 91.13\% & 91.09\% & 91.12\% \\
512           & 90.68\% & 91.19\% & 90.79\% \\
256            & 90.35\% & 90.35\% & 91.19\% \\
128           & 91.12\% & 89.25\% & 89.80\% \\
64            & 85.96\% & 84.98\% & 86.18\% \\
50            & 83.88\% & 82.79\% & 82.35\% \\
40            & 80.15\% & 75.88\% & 81.69\% \\
30            & 74.23\% & 73.90\% & 74.89\% \\
20           & 64.03\% & 58.66\% & 64.58\% \\
10           & 40.90\% & 38.05\% & 41.18\% \\
\bottomrule
\end{tabular}
\end{table}
As can be seen from Table~\ref{tab:comparison}, other point cloud-based classification methods trained on dense point cloud perform poorly with low number of points. Our method with domain adaptation using unlabelled sparse clouds of 50 points helps in increasing classification accuracies even as low as 10 points. 
This further emphasises the fact that popular 3D classification networks are not robust to point density. 
Furthermore, we see that our proposed loss $L_{ours}$ performs favourably in comparison to $L_{CORAL}$ and $L_{MMD}$. 
The marginal improvement may arise from the fact that the dense and sparse dataset is sampled from the same CAD mesh.
We must note that our method involves tuning various hyper-parameters such as $\alpha, \beta$ and $\lambda$ as in Table~\ref{tab:test_alpha},\ref{tab:test_beta} but provides higher flexibility wherein we can provide more importance to source data (in case of noisy target data) or match certain distributional statistics (in case of \textit{a priori} known correlation) between the domains by tuning our hyperparameters. 

\begin{table}[t!]
\centering
\caption{Test set accuracy with $\mathcal{L}_{CORAL}$ over various $\alpha$ $(\lambda=0.5)$}
\label{tab:test_alpha}
\begin{tabular}{cccccc}
\toprule
Points Num & $\alpha=0.1$ & $\alpha=0.5$ & $\alpha=1$ & $\alpha=5$ & $\alpha=10$ \\
\midrule
1024 & 90.35\% & 90.46\% & 89.91\% & 90.57\% & 90.57\% \\
512  & 90.24\% & 91.23\% & 90.24\% & 91.23\% & 90.57\% \\
256  & 89.69\% & 90.24\% & 89.25\% & 90.02\% & 90.35\% \\
128  & 88.05\% & 89.80\% & 87.39\% & 88.60\% & 88.27\% \\
64   & 82.68\% & 84.32\% & 83.11\% & 82.79\% & 85.42\% \\
50   & 81.14\% & 80.04\% & 78.51\% & 78.50\% & 82.02\% \\
40   & 77.30\% & 77.19\% & 75.33\% & 73.14\% & 78.29\% \\
30   & 72.15\% & 70.29\% & 67.98\% & 67.11\% & 72.04\% \\
20   & 58.77\% & 56.80\% & 56.69\% & 56.91\% & 63.60\% \\
10   & 37.28\% & 35.53\% & 32.67\% & 35.75\% & 39.80\% \\
\bottomrule
\end{tabular}
\end{table}

\begin{table}[t!]
\centering
\caption{Test set accuracy with $\mathcal{L}_{ours}$ over various $\beta$ \\$(\alpha=10, \lambda=0.5)$}
\label{tab:test_beta}
\begin{tabular}{ccccc}
\toprule
Points Num & $\beta=0.1$ & $\beta=0.5$ & $\beta=5$  & $\beta=10$ \\
\midrule
1024       & 91.23\%  & 91.12\%  & 90.24\% & 90.13\% \\
512        & 91.67\%  & 90.79\%  & 90.90\% & 90.79\% \\
256        & 90.68\%  & 91.19\%  & 90.46\% & 89.80\% \\
128        & 89.80\%  & 89.80\%  & 88.38\% & 89.04\% \\
64         & 85.11\%  & 86.18\%  & 83.33\% & 84.87\% \\
50         & 82.68\%  & 82.35\%  & 78.73\% & 80.70\% \\
40         & 79.28\%  & 81.69\%  & 75.77\% & 77.52\% \\
30         & 72.92\%  & 74.89\%  & 69.85\% & 69.85\% \\
20         & 59.65\%  & 64.58\%  & 56.69\% & 59.10\% \\
10         & 36.51\%  & 41.18\%  & 37.50\% & 34.43\% \\
\bottomrule
\end{tabular}
\end{table}

\section{Conclusion}
\label{sec:conclusions}
In this work, we tackled the problem of sparse point cloud based object recognition by using unsupervised domain adaptation with dense labelled point clouds and unlabelled sparse point clouds. We proposed a novel domain adaptation loss by combining MMD loss and CORAL loss in a weighted linear combination. We performed extensive experiments on ModelNet10~\cite{wu20153d} with benchmark experiments against state-of-art methods. Our method allows for increased robustness of perceptual pipelines against various types of input data for autonomous agents with minimal overhead costs for data collection and annotation. Future work will focus on extending the proposed methodology to real visuo-tactile data from novel tactile sensors~\cite{ozioko2021sensact, escobedo2020energy, ntagios2020robotic} as well as extend to other contact rich information provided by tactile sensing such as texture, temperature and so on~\cite{kumaresan2021multifunctional,soni2020printed, dahiya2019large, dahiya2019skin}. 

\bibliographystyle{IEEEtran}
\bibliography{IEEEabrv,root}

\begin{thebibliography}{10}
\providecommand{\url}[1]{#1}
\csname url@samestyle\endcsname
\providecommand{\newblock}{\relax}
\providecommand{\bibinfo}[2]{#2}
\providecommand{\BIBentrySTDinterwordspacing}{\spaceskip=0pt\relax}
\providecommand{\BIBentryALTinterwordstretchfactor}{4}
\providecommand{\BIBentryALTinterwordspacing}{\spaceskip=\fontdimen2\font plus
\BIBentryALTinterwordstretchfactor\fontdimen3\font minus
  \fontdimen4\font\relax}
\providecommand{\BIBforeignlanguage}[2]{{%
\expandafter\ifx\csname l@#1\endcsname\relax
\typeout{** WARNING: IEEEtran.bst: No hyphenation pattern has been}%
\typeout{** loaded for the language `#1'. Using the pattern for}%
\typeout{** the default language instead.}%
\else
\language=\csname l@#1\endcsname
\fi
#2}}
\providecommand{\BIBdecl}{\relax}
\BIBdecl

\bibitem{wang2021sparse}
L.~Wang and B.~Goldluecke, ``Sparse-pointnet: See further in autonomous
  vehicles,'' \emph{IEEE Robotics and Automation Letters}, vol.~6, no.~4, pp.
  7049--7056, 2021.

\bibitem{murali2021intelligent}
P.~K. Murali, M.~Kaboli, and R.~Dahiya, ``Intelligent in-vehicle interaction
  technologies,'' \emph{Advanced Intelligent Systems}, p. 2100122, 2021.

\bibitem{Qiang-TRO-2020}
Q.~{Li} \emph{et~al.}, ``A review of tactile information: Perception and action
  through touch,'' \emph{IEEE Trans. on Rob.}, vol.~36, no.~6, pp. 1619--1634,
  2020.

\bibitem{murali2021active}
P.~K. Murali, M.~Gentner, and M.~Kaboli, ``Active visuo-tactile point cloud
  registration for accurate pose estimation of objects in an unknown
  workspace,'' in \emph{2021 IEEE/RSJ International Conference on Intelligent
  Robots and Systems (IROS)}.\hskip 1em plus 0.5em minus 0.4em\relax IEEE,
  2021, pp. 2838--2844.

\bibitem{murali2022active}
P.~K. Murali, A.~Dutta, M.~Gentner, E.~Burdet, R.~Dahiya, and M.~Kaboli,
  ``Active visuo-tactile interactive robotic perception for accurate object
  pose estimation in dense clutter,'' \emph{IEEE Robotics and Automation
  Letters}, vol.~7, no.~2, pp. 4686--4693, 2022.

\bibitem{xiao2021triangle}
C.~Xiao and J.~Wachs, ``Triangle-net: Towards robustness in point cloud
  learning,'' in \emph{Proceedings of the IEEE/CVF Winter Conference on
  Applications of Computer Vision}, 2021, pp. 826--835.

\bibitem{kaboli2019tactile}
M.~Kaboli \emph{et~al.}, ``Tactile-based active object discrimination and
  target object search in an unknown workspace,'' \emph{Autonomous Robots},
  vol.~43, no.~1, pp. 123--152, 2019.

\bibitem{kaboli2018robust}
M.~Kaboli and G.~Cheng, ``Robust tactile descriptors for discriminating objects
  from textural properties via artificial robotic skin,'' \emph{IEEE
  Transactions on Robotics}, vol.~34, no.~4, pp. 985--1003, 2018.

\bibitem{kaboli2018active}
M.~Kaboli, D.~Feng, and G.~Cheng, ``Active tactile transfer learning for object
  discrimination in an unstructured environment using multimodal robotic
  skin,'' \emph{International Journal of Humanoid Robotics}, vol.~15, no.~01,
  p. 1850001, 2018.

\bibitem{kaboli2016tactilehuman}
M.~Kaboli, K.~Yao, and G.~Cheng, ``Tactile-based manipulation of deformable
  objects with dynamic center of mass,'' in \emph{2016 IEEE-RAS 16th
  International Conference on Humanoid Robots (Humanoids)}.\hskip 1em plus
  0.5em minus 0.4em\relax IEEE, 2016, pp. 752--757.

\bibitem{kaboli2017tactile}
M.~Kaboli, D.~Feng, K.~Yao, P.~Lanillos, and G.~Cheng, ``A tactile-based
  framework for active object learning and discrimination using multimodal
  robotic skin,'' \emph{IEEE Robotics and Automation Letters}, vol.~2, no.~4,
  pp. 2143--2150, 2017.

\bibitem{murali2022infogain}
P.~K. Murali, R.~Dahiya, and M.~Kaboli, ``An empirical evaluation of various
  information gain criteria for active tactile action selection for pose
  estimation,'' in \emph{The IEEE Int. Conf on Flexible and Printable Sensors
  and Systems (FLEPS 2022)}.\hskip 1em plus 0.5em minus 0.4em\relax IEEE, 2022.

\bibitem{xiao2021pandaset}
P.~Xiao, Z.~Shao, S.~Hao, Z.~Zhang, X.~Chai, J.~Jiao, Z.~Li, J.~Wu, K.~Sun,
  K.~Jiang \emph{et~al.}, ``Pandaset: Advanced sensor suite dataset for
  autonomous driving,'' in \emph{2021 IEEE International Intelligent
  Transportation Systems Conference (ITSC)}.\hskip 1em plus 0.5em minus
  0.4em\relax IEEE, 2021, pp. 3095--3101.

\bibitem{guo2020deep}
Y.~Guo, H.~Wang, Q.~Hu, H.~Liu, L.~Liu, and M.~Bennamoun, ``Deep learning for
  3d point clouds: A survey,'' \emph{IEEE transactions on pattern analysis and
  machine intelligence}, 2020.

\bibitem{qi2017pointnet}
C.~R. Qi, H.~Su, K.~Mo, and L.~J. Guibas, ``Pointnet: Deep learning on point
  sets for 3d classification and segmentation,'' in \emph{Proceedings of the
  IEEE conference on computer vision and pattern recognition}, 2017, pp.
  652--660.

\bibitem{qi2017pointnet++}
C.~R. Qi, L.~Yi, H.~Su, and L.~J. Guibas, ``Pointnet++: Deep hierarchical
  feature learning on point sets in a metric space,'' \emph{Advances in Neural
  Information Processing Systems}, vol.~30, 2017.

\bibitem{stutz2020learning}
D.~Stutz and A.~Geiger, ``Learning 3d shape completion under weak
  supervision,'' \emph{International Journal of Computer Vision}, vol. 128,
  no.~5, pp. 1162--1181, 2020.

\bibitem{liu2019relation}
Y.~Liu, B.~Fan, S.~Xiang, and C.~Pan, ``Relation-shape convolutional neural
  network for point cloud analysis,'' in \emph{Proceedings of the IEEE/CVF
  Conference on Computer Vision and Pattern Recognition}, 2019, pp. 8895--8904.

\bibitem{wu20153d}
Z.~Wu, S.~Song, A.~Khosla, F.~Yu, L.~Zhang, X.~Tang, and J.~Xiao, ``3d
  shapenets: A deep representation for volumetric shapes,'' in
  \emph{Proceedings of the IEEE conference on computer vision and pattern
  recognition}, 2015, pp. 1912--1920.

\bibitem{wilson2020survey}
G.~Wilson and D.~J. Cook, ``A survey of unsupervised deep domain adaptation,''
  \emph{ACM Transactions on Intelligent Systems and Technology (TIST)},
  vol.~11, no.~5, pp. 1--46, 2020.

\bibitem{wang2018deep}
M.~Wang and W.~Deng, ``Deep visual domain adaptation: A survey,''
  \emph{Neurocomputing}, vol. 312, pp. 135--153, 2018.

\bibitem{borgwardt2006integrating}
K.~M. Borgwardt, A.~Gretton, M.~J. Rasch, H.-P. Kriegel, B.~Sch{\"o}lkopf, and
  A.~J. Smola, ``Integrating structured biological data by kernel maximum mean
  discrepancy,'' \emph{Bioinformatics}, vol.~22, no.~14, pp. e49--e57, 2006.

\bibitem{sun2016return}
B.~Sun, J.~Feng, and K.~Saenko, ``Return of frustratingly easy domain
  adaptation,'' in \emph{Proceedings of the AAAI Conference on Artificial
  Intelligence}, vol.~30, no.~1, 2016.

\bibitem{zhang2018end}
M.~Zhang, Z.~Cui, M.~Neumann, and Y.~Chen, ``An end-to-end deep learning
  architecture for graph classification,'' in \emph{AAAI}, 2018, pp.
  4438--4445.

\bibitem{gretton2012optimal}
A.~Gretton, D.~Sejdinovic, H.~Strathmann, S.~Balakrishnan, M.~Pontil,
  K.~Fukumizu, and B.~K. Sriperumbudur, ``Optimal kernel choice for large-scale
  two-sample tests,'' in \emph{Advances in neural information processing
  systems}.\hskip 1em plus 0.5em minus 0.4em\relax Citeseer, 2012, pp.
  1205--1213.

\bibitem{sun2016deep}
B.~Sun and K.~Saenko, ``Deep coral: Correlation alignment for deep domain
  adaptation,'' in \emph{European conference on computer vision}.\hskip 1em
  plus 0.5em minus 0.4em\relax Springer, 2016, pp. 443--450.

\bibitem{yosinski2014transferable}
J.~Yosinski, J.~Clune, Y.~Bengio, and H.~Lipson, ``How transferable are
  features in deep neural networks?'' \emph{Advances in Neural Information
  Processing Systems}, vol.~27, pp. 3320--3328, 2014.

\bibitem{ozioko2021sensact}
O.~Ozioko, P.~Karipoth, P.~Escobedo, M.~Ntagios, A.~Pullanchiyodan, and
  R.~Dahiya, ``Sensact: The soft and squishy tactile sensor with integrated
  flexible actuator,'' \emph{Advanced Intelligent Systems}, vol.~3, no.~3, p.
  1900145, 2021.

\bibitem{escobedo2020energy}
P.~Escobedo, M.~Ntagios, D.~Shakthivel, W.~T. Navaraj, and R.~Dahiya, ``Energy
  generating electronic skin with intrinsic tactile sensing without touch
  sensors,'' \emph{IEEE Transactions on Robotics}, vol.~37, no.~2, pp.
  683--690, 2020.

\bibitem{ntagios2020robotic}
M.~Ntagios, H.~Nassar, A.~Pullanchiyodan, W.~T. Navaraj, and R.~Dahiya,
  ``Robotic hands with intrinsic tactile sensing via 3d printed soft pressure
  sensors,'' \emph{Advanced Intelligent Systems}, vol.~2, no.~6, p. 1900080,
  2020.

\bibitem{kumaresan2021multifunctional}
Y.~Kumaresan, O.~Ozioko, and R.~Dahiya, ``Multifunctional electronic skin with
  a stack of temperature and pressure sensor arrays,'' \emph{IEEE Sensors
  Journal}, vol.~21, no.~23, pp. 26\,243--26\,251, 2021.

\bibitem{soni2020printed}
M.~Soni, M.~Bhattacharjee, M.~Ntagios, and R.~Dahiya, ``Printed temperature
  sensor based on pedot: Pss-graphene oxide composite,'' \emph{IEEE Sensors
  Journal}, vol.~20, no.~14, pp. 7525--7531, 2020.

\bibitem{dahiya2019large}
R.~Dahiya, N.~Yogeswaran, F.~Liu, L.~Manjakkal, E.~Burdet, V.~Hayward, and
  H.~J{\"o}rntell, ``Large-area soft e-skin: The challenges beyond sensor
  designs,'' \emph{Proceedings of the IEEE}, vol. 107, no.~10, pp. 2016--2033,
  2019.

\bibitem{dahiya2019skin}
R.~Dahiya, ``E-skin: from humanoids to humans,'' \emph{Proceedings of the
  IEEE}, vol. 107, no.~2, pp. 247--252, 2019.

\end{thebibliography}
\end{document}